\documentclass[10pt,journal,compsoc]{./IEEEtran}

\usepackage[utf8]{inputenc} 
\usepackage[T1]{fontenc}    
\usepackage{hyperref}       
\usepackage{url}            
\usepackage{booktabs}       
\usepackage{amsfonts}       
\usepackage{nicefrac}       
\usepackage{microtype}      
\usepackage{color}
\usepackage{graphicx}
\usepackage{subfigure}
\usepackage{float}
\usepackage{amsmath,amssymb}
\usepackage{makecell}

\begin{document}
%
\title{Task-aware Similarity Learning for Event-triggered Time Series}
%
%
%
%

\author{Shaoyu~Dou,
        Kai~Yang,~\IEEEmembership{Senior~Member,~IEEE},
        Yang Jiao,
        Chengbo Qiu,
        and Kui Ren 
\IEEEcompsocitemizethanks{
  \IEEEcompsocthanksitem S. Dou, K. Yang, Y. Jiao and C. Qiu are with the Department of Computer Science
  and Technology, Tongji University, Shanghai 201800, China.\protect\\
  E-mail: kaiyang@tongji.edu.cn \protect\\
}

\thanks{Manuscript received April 19, 2005; revised August 26, 2015.}}

%
%

\markboth{Journal of \LaTeX\ Class Files,~Vol.~14, No.~8, August~2015}%
{Shell \MakeLowercase{\textit{et al.}}: Bare Demo of IEEEtran.cls for Computer Society Journals}

\IEEEtitleabstractindextext{%
\begin{abstract}
Time series analysis has achieved great success in diverse applications such as network security, environmental monitoring, and medical informatics. Learning similarities among different time series is a crucial problem since it serves as the foundation for downstream analysis such as clustering and anomaly detection. It often remains unclear what kind of distance metric is suitable for similarity learning due to the complex temporal dynamics of the time series generated from event-triggered sensing, which is common in diverse applications, including automated driving, interactive healthcare, and smart home automation. The overarching goal of this paper is to develop an unsupervised learning framework that is capable of learning task-aware similarities among unlabeled event-triggered time series. From the machine learning vantage point, the proposed framework harnesses the power of both hierarchical multi-scale sequence autoencoders and Gaussian Mixture Model (GMM) to effectively learn the low-dimensional representations from the time series. Finally, the obtained similarity measure can be easily visualized for explaining. The proposed framework aspires to offer a stepping stone that gives rise to a systematic approach to model and learn similarities among a multitude of event-triggered time series. Through extensive qualitative and quantitative experiments, it is revealed that the proposed method outperforms state-of-the-art methods considerably.
\end{abstract}

}

\maketitle

\IEEEdisplaynontitleabstractindextext

\IEEEpeerreviewmaketitle

\IEEEraisesectionheading{\section{Introduction}\label{sec:introduction}}

\IEEEPARstart{T}{ime}
series is a collection of observed data points listed in time order. Time series analysis plays an essential role in domains as diverse as finance, network security, astronomy, and computer vision. Computing a suitable similarity metric between two time series lies at the heart of a variety of machine learning tasks, such as clustering, anomaly detection, and supervised time series classification. A major application of similarity learning is to detect anomalous or malicious behaviors within a set of Internet of Things (IoT) devices. For instance, a malicious healthcare IoT device may send sensitive personal information to Internet, which compromises users’ privacy and requires immediate attention. As a matter of fact, it has been revealed that numerous IoT devices may send sensitive information of users to unrelated third parties \cite{ren2019information}. More generally, machine learning of time series data such as abnormal IoT behavior detection is arguably expected to be epicenter of a plurality of emerging IoT applications.

\begin{figure}[htbp]
  \centering
  \includegraphics[width=\columnwidth]{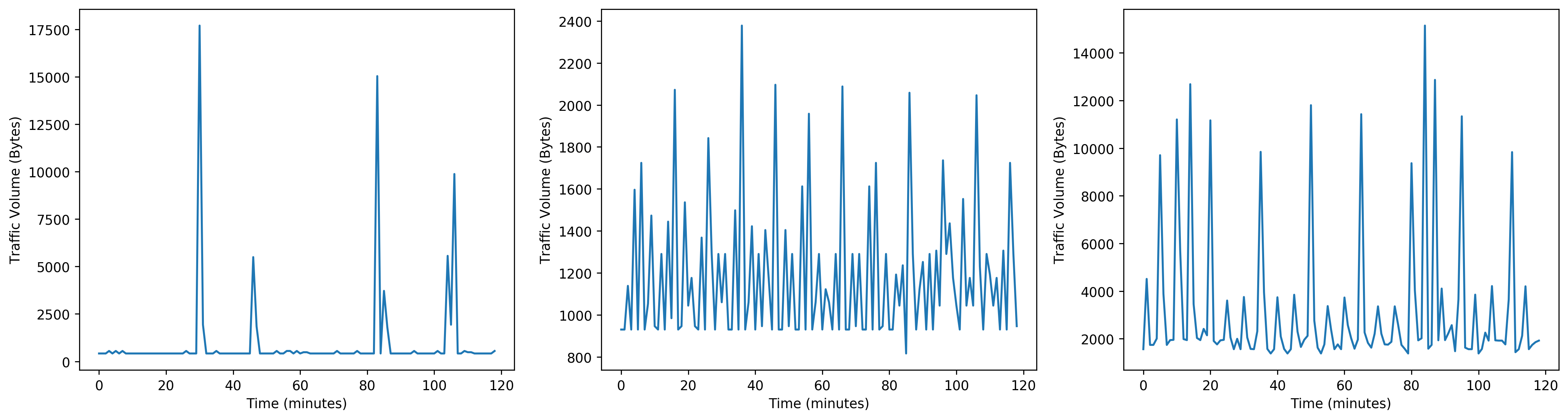}
  \caption{Exemplary event-triggered traffic time series from the UNSW-IoT dataset.}
  \label{fig:example}
\end{figure}

The past decade has witnessed a proliferation in time series data generated from event-triggered sensors, \textcolor{black}{where the events refer to human activities or machine programs. The occurrence of such events will trigger the working state transition of the sensor, resulting in heterogeneous dynamics of the traffic time series. Please refer to Section \ref{sec:event_triggered} for an informal definition of event-triggered time series.} A key challenge in analyzing such event-triggered (traffic) time series is that it often contains temporal event sequence that is sporadic or highly heterogeneous as shown in Figure \ref{fig:example}. It may contain a few short traffic bursts and a long sleep time with no data transmission, as shown in the first subfigure, or seems to be the “superposition” of multiple time series, as shown in the second and third sub-figures. The unique pattern makes this type of time series extremely heterogeneous and exhibit both long-term and short-term temporal dependencies which render the traditional machine learning algorithms not directly applicable. Apart from the heterogeneity, other challenges include: 1) There exists no label for similarity learning. 2) To achieve the best performance, the similarity learning needs to be tuned for a particular task. 3) Time series generated by event-triggered sensors often vary in time granularity across devices due to power and privacy concerns. 4) In many applications, we need to not only compute the similarity between two unlabeled time series but also provide insight into the mechanism so that domain expert can understand the similarity metric. The above challenges give rise to the following questions.

\emph{How can we design a task-aware unsupervised machine learning approach to learn the similarity between two event-triggered time series?}

To this end, we develop a Time Series Hierarchical Multiscale Variational Autoencoder with Gaussian Mixture Model a.k.a. ET-Net to effectively model the temporal dynamics of event-triggered time series. ET-Net leverages both the power of deep autoencoder ensembles to extract multi-span temporal dynamics, and statistical Gaussian Mixture Model (GMM) to measure the similarities among a collection of time series. It can also provide visual outcomes for human understanding. More concretely, we made contributions as listed in the sequel.

\begin{itemize}
\item \textbf{Task-aware unsupervised similarity learning:} ET-Net is completely unsupervised and can learn similarities among unlabeled event-triggered time series without any human intervention. In addition, the similarity learning can be tuned for a particular task to optimize the detection performance.
\item \textbf{Visualization and interpretability:} ET-Net generates a semantically meaningful latent space that can be visualized to explain the learned model. In addition, some classification rules can be drawn from the visualization of the original space to help human experts understand the data.
\item \textbf{Effectiveness:} ET-Net exhibits strong empirical performance for downstream analysis such as clustering and anomaly detection than other competing state-of-the-art methods over real-world datasets. It remains effective even when the training data is contaminated by anomalies or noises that are common in time series analysis \cite{wang2013effectiveness}. In addition, the proposed model still has competitive performance on testing data with different time granularity \cite{eibl2014influence} without retraining.
\end{itemize}



\section{Related Work}
\subsection{Similarity Learning}
We summarize related work on similarity learning that have received significant attentions over the past decade, including non-parametric methods and parametric methods based on deep neural networks.

There exists a large body of work on non-parametric-methods for time series similarity learning, including Euclidean Distance (ED), Editing Distance on Real sequences (EDR) and Dynamic Time Warping (DTW). Euclidean distance, while is the most widely used distance metric, often yields poor performance for time series similarity learning because it is sensitive to anomalies, noise, warping and contamination \cite{toller2019formally}. As a remedy, one may propose to use EDR instead of ED. However, EDR is calculated based on the local procedures and treat all the change operations equally. Therefore, it is less robust to noise, i.e., when a group of data points in the time series deviate slightly, the EDR will become very large. In addition, ED and EDR are sensitive to time series with irregular sampling rates. DTW is a time series similarity metric that has been widely studied and used in recent years. It aims to calculate an optimal matching between two time series and has proved to be capable of providing strong baseline performance in many machine learning tasks such as classification and clustering. In addition, there exists a lot of work dedicated to improving the performance of DTW \cite{silva2016effect}, or combining DTW with deep learning methods \cite{cuturi2017soft, cai2019dtwnet}.

The majority of deep learning-based methods for time series similarity learning use Recurrent Neural Network (RNN) or Convolutional Neural Network (CNN) to model the temporal dynamics and convert them into low-dimensional representations. Similarity or distance metrics in this lowdimensional latent space are expected to reflect semantic relationship between time series. \cite{mathew2019warping} proposes a structure called WaRTEm to generate time series embedding that exhibits resilience to warping. \cite{ma2019learning} proposes a model named DTCR that integrates the seq2seq model and the K-means objective to generate latent space representations that are better suited for clustering. Autowarp proposed in  \cite{abid2018learning} obtains a vector embedding through the sequence autoencoder, which helps to guide the optimization of a warping metric.


\subsection{Visual Interpretability}

Visual interpretability is often seen as the first step in explaining deep neural networks. It can be used to explain the inherent mechanism of the neural network \cite{ma2019learning,zong2018deep,zeiler2014visualizing}, and can also be used to trace which training samples or features significantly affect the output of the neural network, i.e., attributing the output of neural network to a set of features or samples.
In general, the attribution methods can be roughly categorized into two groups, i.e., feature-based and example-based. The feature-based attribution methods compute the contribution of each input feature to the model output, and visualize the result through a heat map superimposed on the input sample. 
\cite{ribeiro2016should} optimizes a white-box model to locally approximate the output of the given neural network, and then determine the contribution of each feature by analyzing the learned white-box model.
\cite{lundberg2017unified} calculates the contribution of each feature based on game theory.
\cite{dabkowski2017real} optimizes a masking model to identify the input features that most influence the decision of the classifier.
\cite{sundararajan2017axiomatic} proposes to use the integrated gradients of the feature as the importance score of the feature.
However, such methods often struggle to generate convincing attribution results on time series data, because the model may highlight features that the model considers important but not to human experts \cite{jeyakumar2020can}.

Example-based attribution methods visualize a set of training samples or prototypes to explain the output of the network. \cite{koh2017understanding} utilizes the influence function to determine which training samples play a decisive role in the model prediction for a given sample. \cite{papernot2018deep} proposes that the K nearest neighbor of a given sample in the feature space are the training samples that contribute the most to the network output. While extensive efforts have been undertaken for the example-based attribution method, its application to time series data is still an under-explored area. 

In this paper, we propose an end-to-end general framework for learning the similarity between event-triggered time series in a fully unsupervised manner. The resulting outcomes can be easily visualized in latent space for human comprehension. We also use the example-based attribution method to explain the model decisions from the original space. Moreover, the proposed model learns the vector embeddings in the latent space by taking into account the machine learning task under investigation. Finally, the probabilistic GMM model offers probabilistic measurement and is more flexible than the K-means clustering method. We summarize the unique features of our model and compare it with other state-of-the-art approaches in Table \ref{table:related}. In particular, the GMM adopted in this framework adopts mixed membership and is much more flexible in terms of cluster covariance than the hard assignment approach such as the K-means clustering method \cite{ma2019learning}. The proposed task-aware approach can learn the vector embeddings that are tailored for a particular machine learning task, so the detection performance can be improved. As evident from the Table \ref{table:related}, only ET-Net meets all the desired requirements.

\begin{table*}[htbp]
  \caption{Comparison of related work}
  \label{table:related}
  \centering
  \scalebox{1}{
  \begin{tabular}{@{}lccccccc@{}}
  \toprule
                            & ED          & DTW         & EDR           & WaRTEm \cite{mathew2019warping} & DTCR \cite{ma2019learning}  & Autowarp \cite{abid2018learning}  & ET-Net     \\ \midrule
  Robustness                & $\times$    & \checkmark  & $\times$      & \checkmark                      & \checkmark                  & \checkmark                        & \checkmark      \\
  Compression               & $\times$    & $\times$    & $\times$      & \checkmark                      & \checkmark                  & \checkmark                        & \checkmark      \\
  Task awareness            & $\times$    & $\times$    & $\times$      & $\times$                        & $\times$                    & $\times$                          & \checkmark      \\
  Flexibility               & $\times$    & $\times$    & $\times$      & $\times$                        & $\times$                    & $\times$                          & \checkmark      \\
  Joint learning            & $\times$    & $\times$    & $\times$      & \checkmark                      & \checkmark                  & $\times$                          & \checkmark      \\ \bottomrule
  \end{tabular}}
\end{table*}
  

\section{Hierarchical Multiscale Variational Autoencoder with Gaussian Mixture Model}
\subsection{Informal Problem Definition}
\label{sec:event_triggered}
\textbf{Event-triggered time series:}
As we enter the era of Internet of Things, more and more time series data are generated from event-triggered sensors. Let ${\underline{\bf{x}}} = [x_1, \cdots, x_L]^T$ denote a time series of length $L$ which is triggered by $K$ events, i.e, 
\begin{equation}
\label{eqn:events}
    {\underline{\bf{x}}} = f({\underline{\bf{a}}}_1 \circ {\underline{\bf{e}}}_1, \cdots, {\underline{\bf{a}}}_K \circ {\underline{\bf{e}}}_K),
\end{equation}
where ${\underline{\bf{e}}}_i$ represents a time indicator sequence of length $L$, ${\underline{\bf{a}}}_i$ represents corresponding intensity vector of length $L$, and $\circ$ is Hadamard product.
Traffic time series in IoT network are usually caused by two types of events: machine type communication (MTC) events, which trigger short-term and periodic dependency, and human type communication (HTC) events, which trigger long-term and bursty dependency \cite{tahaei2020rise}.

Let's consider the traffic generated by a webcam. Its traffic in a given time window is equal to the sum of traffic generated by all events during the period. i.e.,  ${\underline{\bf{x}}} = {\underline{\bf{a}}}_1 \circ {\underline{\bf{e}}}_1 + \cdots + {\underline{\bf{a}}}_K \circ {\underline{\bf{e}}}_K$, where ${\underline{\bf{a}}}_i$ is the traffic generated by type-$i$ event. 
The interaction events between the camera and the controller, such as routine queries and responses, are MTC events. In a given time window, these events occur frequently but consume less traffic, so they can be regarded as background traffic in the traffic time series, which constitutes short-term dependencies in the time series.
In contrast, human interactions such as viewing surveillance video, are HTC events. In the above time window, these events occur much less frequently than MTC events, but they consume a lot of traffic and appear as bursts in the traffic time series.
We summarize the characteristics of MTC and HTC events in Table \ref{table:mtc_htc}.
     
\begin{table}[htbp]
    \caption{Characteristics of MTC and HTC events}
    \label{table:mtc_htc}
    \centering
    \begin{tabular}{@{}ccl@{}}
        \toprule
        Event Type & Volume & \multicolumn{1}{c}{Pattern}                                                                             \\ \midrule
        MTC        & Small  & \begin{tabular}[c]{@{}l@{}}Almost periodic\\ Short transmission period\end{tabular}                     \\
        HTC        & Large  & \begin{tabular}[c]{@{}l@{}}Bursty and unpredictable\\ Generally long transmission interval\end{tabular} \\ \bottomrule
    \end{tabular}
\end{table}
  
Given a collection of event-triggered time series ${\bf{X}} = \left[ {\underline{\bf{x}}}_1, {\underline{\bf{x}}}_2 , \cdots , {\underline{\bf{x}}}_N\right]$. We aim to output a low-dimensional representation for each time series to compute the similarities between the time series.

\subsection{General Framework}
As discussed in the last chapter, the heterogeneity in IoT traffic mainly stems from two types of data traffic, i.e., traffic caused by HTC and MTC events. MTC events typically span a short period of time and exhibit periodic behavior. On the contrary, HTC events give rise to long-term and bursty traffic which are more difficult to characterize and predict. To fully characterize heterogeneous IoT traffic caused by both MTC and HTC events, we present in this chapter the ET-Net framework.

The proposed framework for time series similarity learning is composed of two modules: compression network and distribution estimator. The former compresses time series into latent space ${\cal Z}$, and the latter estimates the latent space distribution. These two modules work in a coordinated manner to jointly learn the temporal dynamics of the event-triggered time series and generate vector embeddings tailored for GMM.
The objective function for learning this model is given in the sequel.
\begin{equation}
\label{general_eqn:loss}
    L = \left\| {{\bf{X}} - g({\bf{X}})} \right\|_2 ^2  + \lambda E({\bf{Z}}, {\bf{X}}),
\end{equation}
where $g(\cdot)$ is the reconstruction model. $\bf{Z}$ is the representation of $\bf{X}$ in the latent space. \textcolor{black}{$\left\|  \cdot  \right\|_2 ^2$ denotes the squared L2-loss.}
\textcolor{black}{$E(\cdot)$ is the negative log-likelihood of the estimated GMM, a.k.a an energy function, which models latent space distribution.}
$\lambda$ is a weighting parameter that governs the tradeoff between two individual objective functions.

The above formula is similar to a variety of existing work, including VAE \cite{kingma2013auto}, DAGMM \cite{zong2018deep}, DTCR \cite{ma2019learning} and SOM-VAE \cite{fortuin2018som}. In this paper, we outline a special type of sequence autoencoder architecture that is particularly suitable for learning similarity among event-triggered time series.

\subsection{Model Overview}
We construct a compression network using a structure partly similar to seq2seq \cite{sutskever2014sequence}, which includes an encoder and a decoder, i.e., ${\underline{\bf{z}_c}}  = g_e ({\underline{\bf{x}}})$, ${{\underline{\bf{x}}}}' = g_d ({\underline{{\bf{z}_c}}})$, where $g_e (\cdot)$ and $g_d(\cdot)$ denote encoder function and decoder function, respectively. ${\underline{\bf{z}_c}}$ represents the compressed latent space representation of the time series ${\underline{\bf{x}}}$. ${{\underline{\bf{x}}}}'$ denotes the reconstructed time series.

We further stack the compressed representation with the reconstruction errors to obtain the \textit{extended latent space} representation ${\underline{\bf{z}}}$,
\begin{equation}
\label{eqn:1}
    {\underline{\bf{z}}} = \left[ {\underline{{\bf{z}_c}} ,d_{rel}({\underline{\bf{x}}}',{\underline{\bf{x}}}), d_{cos}({\underline{\bf{x}}}',{\underline{\bf{x}}})} \right],
\end{equation}
where $d(\cdot)_{rel}$ and $d(\cdot)_{cos}$ are the reconstruction error, denoting the relative distance and cosine similarity, respectively.
Once the extended latent space representation is obtained, it is fed into a GMM estimator for density estimation, as shown in the sequel.
\begin{equation}
\label{eqn:2}
    \begin{array}{l}
        {\underline{\boldsymbol{\gamma}}} = g_m\left( {\underline{\bf{z}}} \right)\quad
        \varphi _k  = \sum\nolimits_{i = 1}^M {\frac{{\boldsymbol{\gamma} _{ik} }}{M}}  \quad
        \mu _k ,\Sigma _k  = \eta \left( {\{ [{\underline{\bf{z}}}_i,{\underline{\boldsymbol{\gamma}}}_i ]\} _{i = 1}^M } \right),
    \end{array}
\end{equation}
where $K$ is the number of mixture components in GMM and $M$ is the number of samples in mixture component $k$. $g_m(\cdot)$ is a membership estimator, and ${\underline{\boldsymbol{\gamma}}}$ is a $K$-dimensional vector representing the probability that sample ${\underline{\bf{z}}}$ belonging to the $k^{th}$ mixture component.
$\varphi _k$, $\mu _k$ and $\Sigma _k$ are mixture probability, mean and covariance for $k^{th}$ mixture component, respectively.
$\eta(\cdot)$ denotes a function for computing the mean and covariance.
In practice, we use the iterative EM algorithm to update $\mu _k$ and $\Sigma _k$ based on ${\underline{\bf{z}}}$, instead of computing them directly using ${\underline{\boldsymbol{\gamma}}}$ and ${\underline{\boldsymbol{\varphi}}}$ like DAGMM.

Once we obtain the parameters of the GMM model, the sample energy function (c.f. (\ref{general_eqn:loss})) can be calculated as follows,
\begin{equation}
\label{eqn:energy}
    \begin{array}{l}
        E ({\underline{\bf{z}}} , {\underline{\bf{x}}}) =  - \log \left( {\sum\limits_{k = 1}^K {\varphi _k \theta \left( {{\underline{\bf{z}}}|(\mu _k ,\Sigma _k )} \right)} } \right) \\
        \theta \left( {{\underline{\bf{z}}}, (\mu _k ,\Sigma _k )} \right) = \frac{1}{{\sqrt {\left| {2\pi \Sigma _k } \right|} }}\exp \left( { - \frac{{({\underline{\bf{z}}} - \mu _k )^2 }}{{2\Sigma _k }}} \right)
    \end{array}
\end{equation}
where the $ \theta \left( {{\underline{\bf{z}}}, (\mu _k ,\Sigma _k )} \right)$ is the density function of Gaussian distribution ${\cal N}(\mu _k ,\Sigma _k )$. The overall loss function is given below,
where $N$ is the number of training samples.

\begin{equation}
\label{eqn:loss}
    L = {1 \over N} \sum\limits_{i = 1}^N {\left\| {{\underline{\bf{x}}}'_i - {\underline{\bf{x}}}_i} \right\|_2 ^2 }  + {\lambda \over N} \sum\limits_{i = 1}^N {E({\underline{\bf{z}}}_i , {\underline{\bf{x}}}_i)}.
\end{equation}

However, due to the temporal heterogeneity, complex long- and short-term dependencies of event-triggered time series make traditional sequence autoencoder ineffective in capturing time dynamics.

\begin{figure*}
\centering
    \includegraphics[width=1.8\columnwidth]{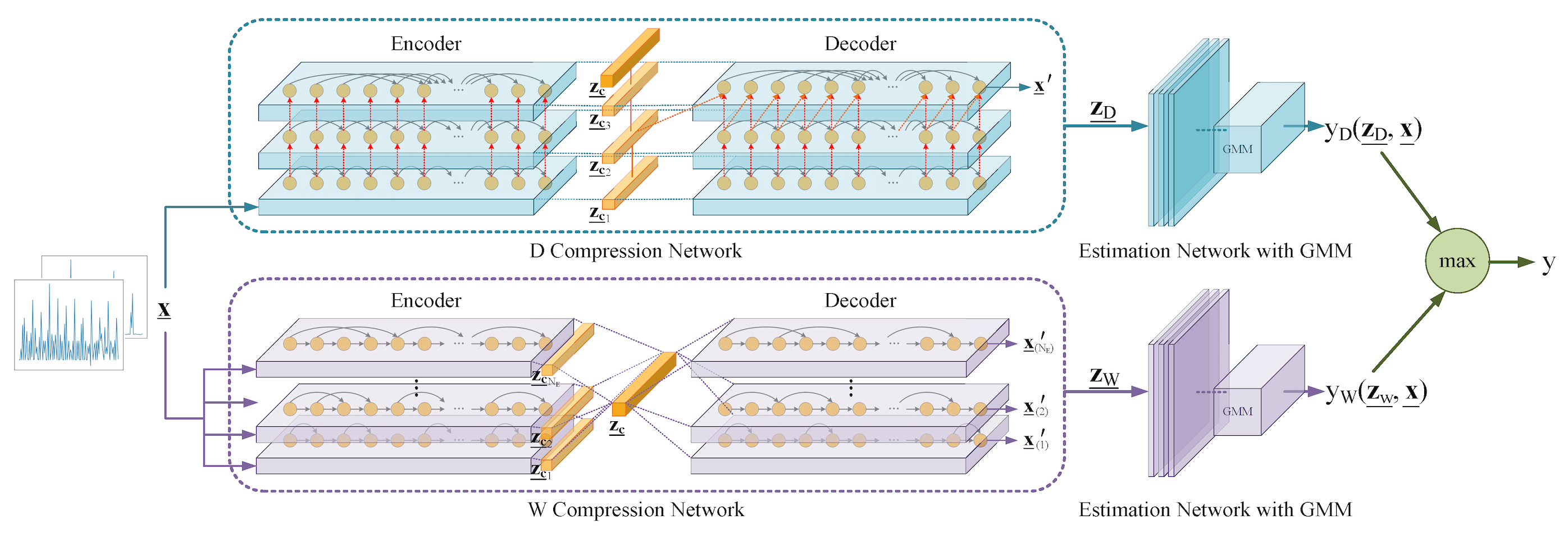}
    \caption{The architecture of ET-Net}
\label{fig:framework}
\end{figure*}

To this end, we propose the ET-Net, as shown in Figure \ref{fig:framework}. Specifically, we propose W and D compression networks to extract the features of event-triggered time series, which give rise to low-dimensional representations. The GMM model will subsequently estimate the distribution in the latent space and output learning outcomes, i.e. $y_W$ and $y_D$.

\subsubsection{W compression network}

\textcolor{black}{The intuition behind creating a W compression network is to learn the temporal dependencies between various bursts that may represent HTCs. Each branch of the encoder-decoder pair randomly learns a type of dependencies with a different temporal span. All learned representations are compressed into a latent space representation, which is borrowing from the idea of multi-task learning \cite{long2017learning}.}
Figure \ref{fig:framework} shows the architecture of W compression network.
Reconstructed sequence output by $i^{th}$ decoder ${\underline{\bf{x}}}'_{(i)}$ is computed as
${\underline{\bf{z_{c}}}_i}  = g_{ei} ({\underline{\bf{x}}})$ and ${\underline{\bf{x}}}'_{(i)}  = g_{di} ({\underline{\bf{z_{c}}}})$,
where ${\underline{\bf{z_{c}}}_i}$ is the final state of $i^{th}$ encoder. The final state of all encoder branches are compressed into a latent space representation ${\underline{{\bf{z}_c}}}$, ${\underline{{\bf{z}_c}}}  = {\bf{W}}^{W}  \cdot [{\underline{\bf{z_{c}}}_1} , \cdots ,{\underline{\bf{z_{c}}}_{N_E}}]^T + {\underline{\bf{b}}}^{W}$,
where $N_E$ is the number of encoders/decoders, ${\bf{W}}^{W}$ and
${\underline{\bf{b}}}^{W}$ denote a trainable weight matrix and a bias vector,
respectively.

The \textit{recurrent function} used to update the hidden state of the recurrent cell in the $i$-th layer of SRNN \cite{kieu2019outlier} is as follows,
\begin{equation}
\label{eqn:srnn}
    \begin{array}{l}
        {\underline{{\bf{h}}}}^i (t)  = \frac{ w_1^i(t) \cdot f_{rnn} ({\underline{{\bf{h}}}}^i (t-1) ,x(t) ) + w_2^i(t)  \cdot f'({\underline{{\bf{h}}}}^i(t-s^i) ,x(t) )}{w_1^i(t) + w_2^i(t)} \\
        s.t.\quad w_1^i(t) ,w_2^i(t)  \in \{ 0,1\} , w_1^i(t)  + w_2^i(t)  \ne 0 \\
    \end{array}
\end{equation}
where $w_1^i(t)$ and $w_2^i(t)$ are randomly initialized weights.
$f_{rnn} (\cdot)$ denotes a non-linear function including Long-Short Term Memory (LSTM)
\cite{hochreiter1997long} or Gated Recurrent Unit (GRU) \cite{chung2014empirical}.
$f'(\cdot)$ denotes a linear operation.
$s^i$ is a parameter that controls the memory ability of SRNN.
When $s^i$ is small, SRNN tends to learn short-term dependencies.
Otherwise, it will learn long-term dependencies.
We set the parameter $s^i$ of each encoder/decoder to a different value but no more than three, so that it tends to learn short-term dependencies in time series.

When LSTM is set as the recurrent cell, $f_{rnn}(\cdot)$ in (\ref{eqn:srnn}) can be expanded as $f_{lstm}(\cdot)$,
\begin{equation}
\label{eqn2}
    \begin{array}{l}
        f_{lstm} ({\underline{{\bf{h}}}}(t - 1),{\underline{{\bf{c}}}}(t - 1),x(t)) = {\underline{{\bf{o}}}}(t) \circ  {\underline{{\bf{c}}}}(t) \\
        {\underline{{\bf{o}}}}(t) = \sigma ({\bf{W}}_o  \cdot [{\underline{{\bf{h}}}}(t - 1) , x(t)] + {\underline{{\bf{b}}}}_o ) \\
        {\underline{{\bf{c}}}}(t) = {\underline{{\bf{f}}}}(t) \circ {\underline{{\bf{c}}}}(t - 1) + {\underline{{\bf{i}}}}(t) \circ {\underline{{\bf{\tilde c}}}}(t) \\
        {\underline{{\bf{f}}}}(t) = \sigma ({\bf{W}}_f  \cdot [{\underline{{\bf{h}}}}(t - 1) , x(t)] + {\underline{{\bf{b}}}}_f ) \\ 
        {\underline{{\bf{i}}}}(t) = \sigma ({\bf{W}}_i  \cdot [{\underline{{\bf{h}}}}(t - 1) , x(t)] + {\underline{{\bf{b}}}}_i ) \\
        {\underline{{\bf{\tilde c}}}}(t) = \tanh ({\bf{W}}_c  \cdot [{\underline{{\bf{h}}}}(t - 1) , x(t)] + {\underline{{\bf{b}}}}_c )
    \end{array}
\end{equation}  
where ${\underline{{\bf{i}}}}$, ${\underline{{\bf{o}}}}$ and ${\underline{{\bf{f}}}}$ are input gate, output gate and forget gate respectively. ${\underline{{\bf{c}}}}$ is the memory.
${\bf{W}}_o$, ${\bf{W}}_f$, ${\bf{W}}_i$, ${\bf{W}}_c$, ${\underline{{\bf{b}}}}_o$, ${\underline{{\bf{b}}}}_f$, ${\underline{{\bf{b}}}}_i$ and ${\underline{{\bf{b}}}}_c$ are the parameters to be learned. The $f_{rnn}(\cdot)$ is defined as $f_{gru}(\cdot)$ when GRU is set as the recurrent cell.
\begin{equation}
\label{eqn4}
    \begin{array}{l}
        f_{gru} ({\underline{{\bf{h}}}}(t - 1),x(t)) = ({\underline{{\bf{1}}}} - {\underline{{\bf{u}}}}(t)) \circ {\underline{{\bf{h}}}}(t - 1) + {\underline{{\bf{u}}}}(t) \circ {\underline{{\bf{ \tilde h}}}}(t) \\
        {\underline{{\bf{u}}}}(t) = \sigma ({\bf{W}}_u  \cdot [{\underline{{\bf{h}}}}(t - 1) , x(t)] + {\underline{{\bf{b}}}}_u) \\
        {\underline{{\bf{ \tilde h}}}}(t) = \tanh ({\bf{W}}_h  \cdot [({\underline{{\bf{r}}}}(t) \circ {\underline{{\bf{h}}}}(t - 1)) ,  x(t)] + {\underline{{\bf{b}}}}_h) \\
        {\underline{{\bf{r}}}}(t) = \sigma ({\bf{W}}_r  \cdot [{\underline{{\bf{h}}}}(t - 1) , x(t)] + {\underline{{\bf{b}}}}_r)
    \end{array}
\end{equation} 
where ${\underline{{\bf{u}}}}$ and ${\underline{{\bf{r}}}}$ are update gate and reset gate respectively,
${\bf{W}}_u$, ${\bf{W}}_h$, ${\bf{W}}_r$, ${\underline{{\bf{b}}}}_u$, ${\underline{{\bf{b}}}}_h$ and ${\underline{{\bf{b}}}}_r$ are parameters to be learned.
    
The extended latent space representation in (\ref{eqn:1}) is then given by
${\underline{\bf{z_W}}} = [{\underline{\bf{z}_c}} , d_{rel}({\underline{\bf{x}}},{\underline{\bf{x}}}'_{(i)}), d_{cos} ({\underline{\bf{x}}},{\underline{\bf{x}}}'_{(j)})]$
in W compression network, where $i$ and $j$ are the autoencoder branch indexes with the minimum reconstruction relative distance and cosine distance, respectively. Consequently, we obtain the following loss function for W compression network and the associated GMM.
\begin{equation}
\label{eqn:l_wider}
    L = \frac{1}{{N N_E }} \sum\limits_{i = 1}^N {\sum\limits_{j = 1}^{N_E} { \left\| {{\underline{\bf{x}}}_i - {\underline{\bf{x}}}_i {'}_{(j)}}  \right\|_2 ^2 }} + \frac{\lambda}{{N}} \sum\limits_{i = 1}^N {E({\underline{\bf{z}}}_i , {\underline{\bf{x}}}_i)}
\end{equation}

\subsubsection{D compression Network}
\textcolor{black}{The intuition behind creating a D compression network is to learn the long-term background dynamic of IoT time series that may represent MTCs. Multiple layers of dilated RNNs are stacked sequentially to obtain deep representations of time series.}
Figure \ref{fig:framework} shows the architecture of D compression network,
The latent space representation ${\underline{\bf{z_c}}}$ is computed as
${\underline{\bf{z_c}}} = {\bf{W}}^{D}  \cdot [{\underline{\bf{z_{c}}}_1} , \cdots ,{\underline{\bf{z_{c}}}_{N_L}}]^T + {\underline{\bf{b}}^{D}}$, 
where ${\underline{\bf{z_c}}}_i$ represents the final state of $i^{th}$ layer. $N_L$ is the number of layers in encoder and decoder. ${\bf{W}}^{D}$ and ${\underline{\bf{b}}^{D}}$ denote a trainable weight matrix and a bias vector, respectively.

The hidden state of the recurrent cell in the $i$-th layer of dilated RNN is updated as
\begin{equation}
\label{eqn:drnn}
    \begin{array}{l}
        {\underline{\bf{h}}}^i (t) = f({\underline{\bf{h}}}^{i-1} (t) ,{\underline{\bf{h}}}^i (t-d^i) ) \\
        {\underline{\bf{h}}}^0 (t) = x(t)
    \end{array}
\end{equation}
where $d^i$ denotes the dilation size in $i^{th}$ layer. The hidden state at time instance $t$ only depends on the state at $t-d^i$. Thus, $d^i$ governs the time scale of the dependency that the network aims to mine. In addition, a multi-layer dilated RNNs are used to further extract long-term dependencies by stacking multiple layers with different dilations. In practice, we set 3 as the dilations in first layer, then an exponential growth strategy is used to set the dilation in subsequent layer, that is, $d^i = 3^{i-1}$. We then generate the extended latent space representation as ${\underline{\bf{z_D}}} = [{\underline{\bf{z_c}}}, d_{rel} ({\underline{\bf{x}}}', {\underline{\bf{x}}}), d_{cos} ({\underline{\bf{x}}}', {\underline{\bf{x}}})]$, which is similar to (\ref{eqn:1}). The loss function for D compression network remains the same to (\ref{eqn:loss}).

Please note that the final output \textcolor{black}{of the ET-Net} $y$ is calculated according to the machine learning task we aim to carry out. For anomaly detection, $y$ represents the anomaly score of the sample ${\underline{\bf{x}}}$, and is calculated as $y = \max (E_{W} ({\underline{\bf{z}_W}}, {\underline{\bf{x}}}),E_{D} ({\underline{\bf{z}_D}}, {\underline{\bf{x}}}))$, where $E_{W}(\cdot)$ and $E_{D} (\cdot)$ denote the energy functions of W and D branches, respectively. \textcolor{black}{In this way, the W and D branches identify anomalies from the perspective of HTC and MTC respectively, and the network outputs the highest anomaly score to ensure high recall.}
For clustering or classification task, $y$ represents the predicted label, and defined as $y = \arg \max _i ( {\max ( {[ {{\underline{\boldsymbol{\gamma}_W}},{\underline{\boldsymbol{\gamma}_D}}} ]^T } )} )$, where ${\underline{\boldsymbol{\gamma}_W}}$ and ${\underline{\boldsymbol{\gamma}_D}}$ represent \textcolor{black}{probabilistic GMM membership} predicted by W and D branches respectively. \textcolor{black}{Here we take the maximum of the two predicted probabilities, the result output by the sharper softmax distribution are preferred \cite{hendrycks2016baseline}.}

\section{Experiments}
Two machine learning tasks have been considered in the experiment, i.e.,
anomaly detection and clustering. We carry out the study to answer the following
questions regarding the proposed approach. 1) Effectiveness: whether ET-Net
outperforms the existing state-of-the-art anomaly detection and clustering methods?
2) Robustness: is ET-Net robust to noise and training sample contamination? Whether the trained model is capable of being robust to time granularity variations, which often occurs during practical deployment?
3) Visualization: can we visualize and interpret the similarity metric learned by ET-Net?

\subsection{Datasets and Experimental Design}
\subsubsection{Datasets} For anomaly detection, we first conduct experiments on a synthetic dataset. Then we conduct experiments on several public real-world traffic datasets, i.e., 
UNSW-IoT Dataset\footnote{\url{https://iotanalytics.unsw.edu.au/iottraces.html}} \cite{sivanathan2018classifying},
cell traffic dataset\footnote{\url{https://dandelion.eu/datagems/SpazioDati/telecom-sms-call-internet-mi}},
IoT23 dataset\footnote{\url{https://www.stratosphereips.org/datasets-iot23}},
and datasets selected from UCR time series classification archives\footnote{\url{https://www.cs.ucr.edu/~eamonn/time_series_data_2018}}.
AUC (Area under the Receiver Operating Curve) is employed to assess the
anomaly detection performance.

In UNSW-IoT and IoT23 dataset, packets generated by all devices are captured and recorded,
and each device is identified by a unique label.
We split the entire traffic time series into non-overlapping windows, each
spanning a time interval of one hundred and twenty minutes. Each data point
within the time window represents the number of packets collected within
one minute.
For UNSW-IoT dataset, we aim to detect non-IoT devices within an IoT network since
these non-IoT devices need to be managed and secured using policies different
from that of the IoT devices. \cite{ortiz2019devicemien,sivanathan2017characterizing}.
For IoT23 dataset, we aim to detect malicious attacks in an IoT network.
The cell traffic dataset is generated from the Call Detail Record (CDR), and
the time series from a selected cell is deemed as non-anomalous data. We
then inject the traffic time series from another cell to create anomalies.
For UNSW-IoT, IoT23 and cell traffic datasets, we divide them into training
and test datasets with a 40-60 split. For UCR dataset, we follow the method
described in \cite{benkabou2018unsupervised} to conduct the anomaly detection
experiments.

Likewise, the clustering pxerformance of the proposed framework is also elucidated
via experiments on synthetic datasets and three real-world datasets,
including UNSW-IoT, IoT23 and cell traffic.
NMI (Normalized Mutual Information) is employed to assess the clustering
performance.

\subsubsection{Baselines}
For anomaly detection, we compare ET-Net against the following state-of-the-art unsupervised methods, including One-Class SVM (OCSVM),
Local Outlier Factor (LoF), Isolation Forest (IF), Dynamic Time Warping (DTW), GRU-AE \cite{malhotra2016lstm}, Shared-SRNN \cite{kieu2019outlier}, DAGMM \cite{zong2018deep}, and BeatGAN \cite{zhou2019beatgan}.

Similarly, the baseline algorithms for clustering include K-means, GMM, K-means+DTW, K-means+EDR \cite{chen2005robust}, K-shape \cite{paparrizos2015k}, DEC \cite{xie2016unsupervised}, IDEC \cite{guo2017improved}, SPIRAL \cite{lei2019similarity}, DTC \cite{madiraju2018deep}, and Autowarp \cite{abid2018learning}.

More details about the experiments can be found in the supplementary materials.
For fair comparison, all baseline methods use the parameter settings recommended by authors.

\subsection{Latent Space Visualization on Synthetic Datasets}
In this section, we visualize the latent space learned by ET-Net on anomaly detection and clustering tasks to elucidate the underlying mechanism of the ET-Net, and provide an intuitive explanation for the model outcomes. Finally, we examine the robustness of latent space representation against data granularity variations and different types of noise.

\subsubsection{Anomaly detection}
\begin{figure}[htbp]
\centering
\subfigure[Sine wave]{
\begin{minipage}{0.31\columnwidth}
    \includegraphics[width=\columnwidth]{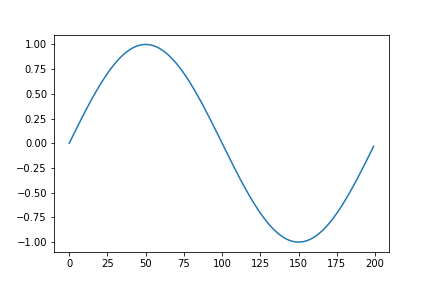}
\end{minipage}}
\subfigure[Square wave]{
\begin{minipage}{0.31\columnwidth}
    \includegraphics[width=\columnwidth]{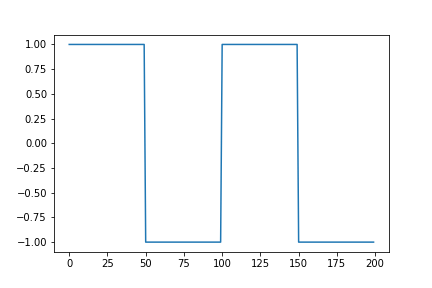}
\end{minipage}}
\subfigure[Triangle wave]{
\begin{minipage}{0.31\columnwidth}
    \includegraphics[width=\columnwidth]{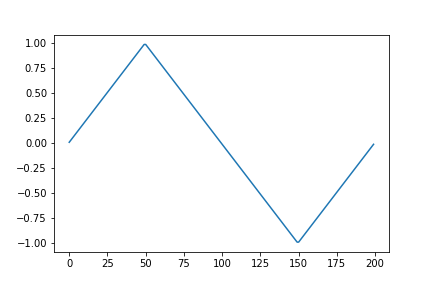}
\end{minipage}}

\caption{Exemplary samples from the synthetic dataset}
\label{fig:exp_anomaly}
\end{figure}

We first assess the performance of the proposed framework via conducting machine learning tasks on a synthetic dataset, as shown in Figure \ref{fig:exp_anomaly}. This dataset consists of three non-anomalous time series samples, i.e., a sine wave, a square wave, and a triangle wave. We also create a total of five hundred copies for each time series and use them to train the proposed deep learning model.

\begin{figure*}[!t]
\centering
    \includegraphics[width=1.8\columnwidth]{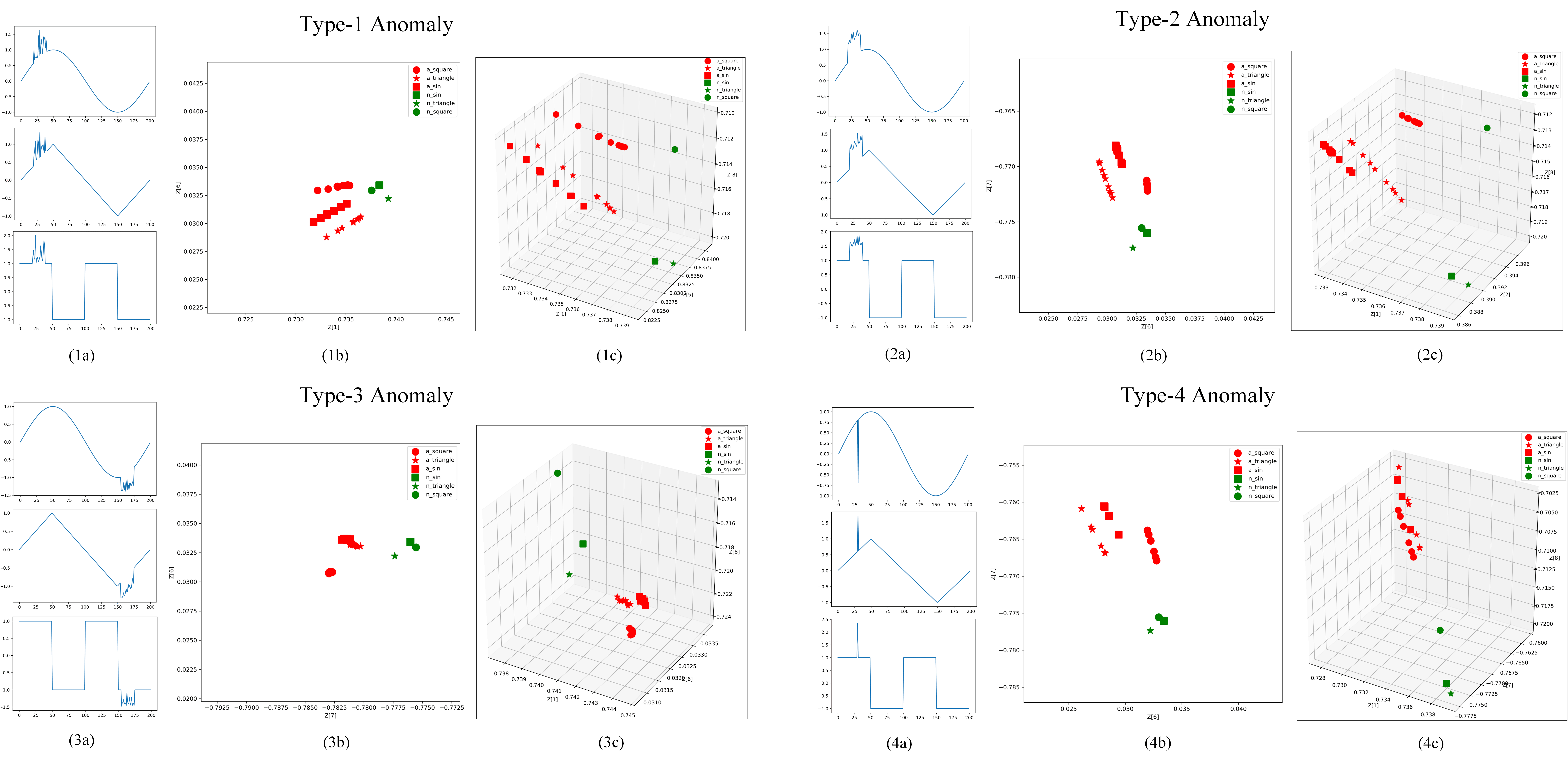}
    \caption{
    (1a)-(4a) Exemplary samples of type-1 to type-4 anomaly.
    (1b)-(4b) 2D visualization of latent space representations of non-anomalous time series (green symbols) and anomaly time series (red symbols).
    (1c)-(4c) 3D visualization of latent space representations of non-anomalous time series (green symbols) and anomaly time series (red symbols).}
\label{fig:result_anomaly}
\end{figure*}

A total of four types of anomalies\footnote{\url{https://anomaly.io/anomaly-detection-twitter-r/}} have been generated to assess the effectiveness of the proposed ET-Net framework. 
Type-1 anomaly refers to strong local additive white Gaussian noise, as shown in Figure \ref{fig:result_anomaly}(1a).
Type-2 anomaly stands for an unusually high activity that spans a short period of time (Figure \ref{fig:result_anomaly}(2a)).
Type-3 is the ``breakdown'' anomaly (Figure \ref{fig:result_anomaly}(3a)).
Type-4 anomaly is created by adding an impulse noise into the time series, as shown in Figure \ref{fig:result_anomaly}(4a).
Both the time domain and latent space representations are illustrated. In the second and third sub-figures of Figure \ref{fig:result_anomaly}, the green symbols represent non-anomalous time series while the red ones correspond to anomaly time series. It is seen through both 2D and 3D visualization that ET-Net can effectively separate the anomalous time series from non-anomalous ones in the latent space.

\begin{figure}[htbp]
\centering
    \includegraphics[width=0.52\columnwidth]{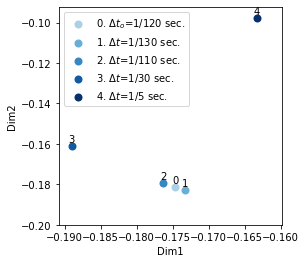}
    \caption{Latent space representation of sine signals with different sampling intervals.}
\label{fig:latent_sin_diff}
\end{figure}

\begin{figure*}[!t]
\centering
    \includegraphics[width=1.8\columnwidth]{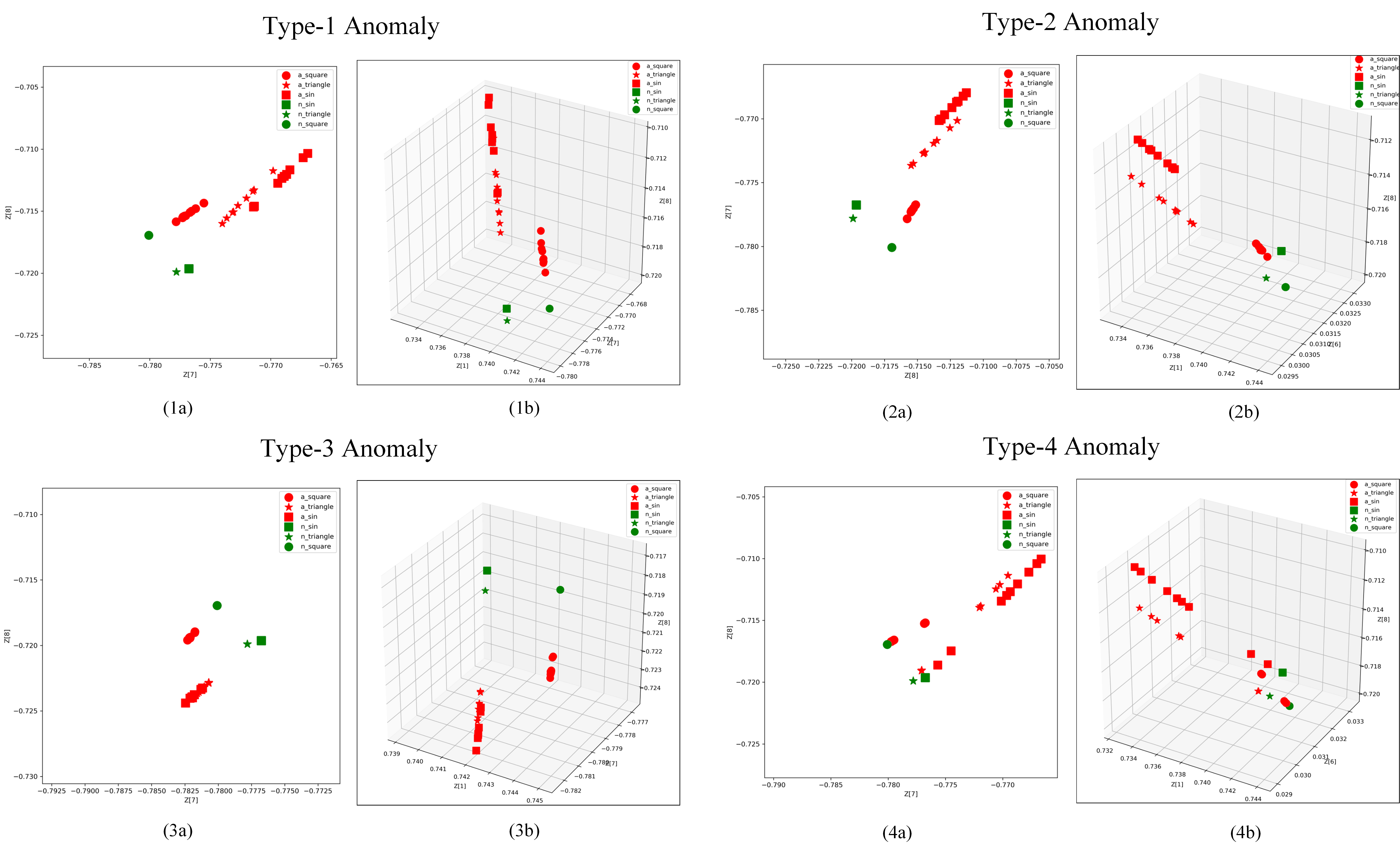}
    \caption{Latent space representations of type-1 to type-4 anomalies with different
    sampling intervals. Both non-anomalous time series (green symbols) and anomaly
    time series (red symbols) are shown in the figure.
    (1a)-(4a) 2D visualization.
    (1b)-(4b) 3D visualization.}
\label{fig:c_result_anomaly}
\end{figure*}

\textbf{Robustness against time granularity variations:} We first evaluate whether the latent space representations remain robust when the data granularity changes. Taking a sine signal (Figure \ref{fig:exp_anomaly}(a)) as an example. We set the original sampling interval $\Delta t_o$ to $1/120$ second, and then vary the sampling rates from $\Delta t = 1/110, 1/130$ second to $\Delta t = 1/30, 1/5$ second, and visualize the obtained latent space representations in Figure \ref{fig:latent_sin_diff}. The visualization result shows that the latent space representation generated by ET-Net is robust against variations in time granularity.
Next we apply the obtained model to a test time series dataset in which the sampling rate is different from that of the training dataset.
Figure \ref{fig:c_result_anomaly} illustrates the latent space representations of type-1 to type-4 anomaly time series and corresponding non-anomalous time series, where red and green symbols represent abnormal and normal time series, respectively. It is seen that a machine learning model trained by the proposed method can be applied to time series with a different time granularity without any model retraining.

\subsubsection{Clustering}
\begin{figure}[htbp]
\centering
    \subfigure[Sine waves]{
    \begin{minipage}{0.31\columnwidth}
        \includegraphics[width=\columnwidth]{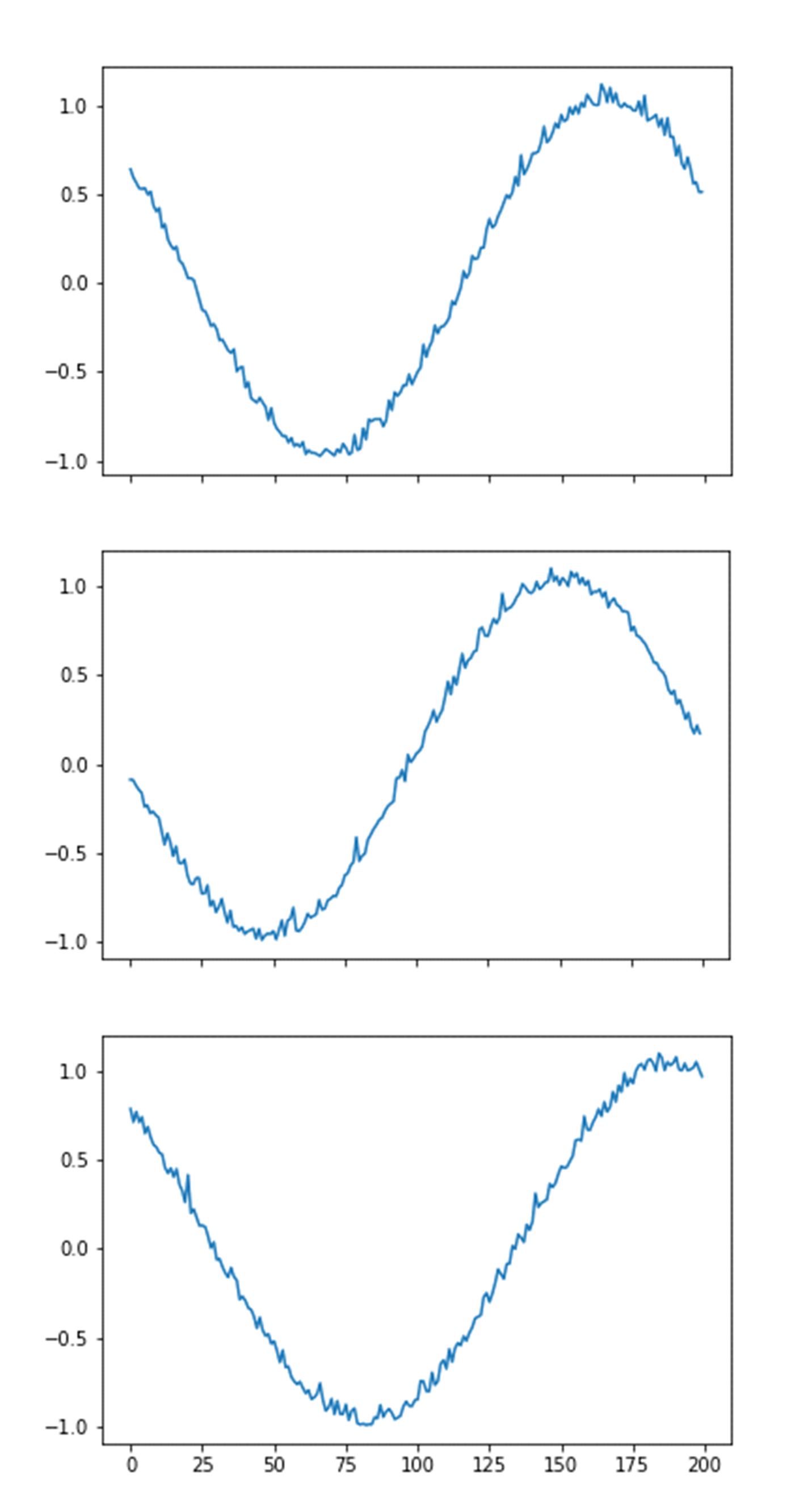}
    \end{minipage}}
    \subfigure[Square waves]{
    \begin{minipage}{0.31\columnwidth}
        \includegraphics[width=\columnwidth]{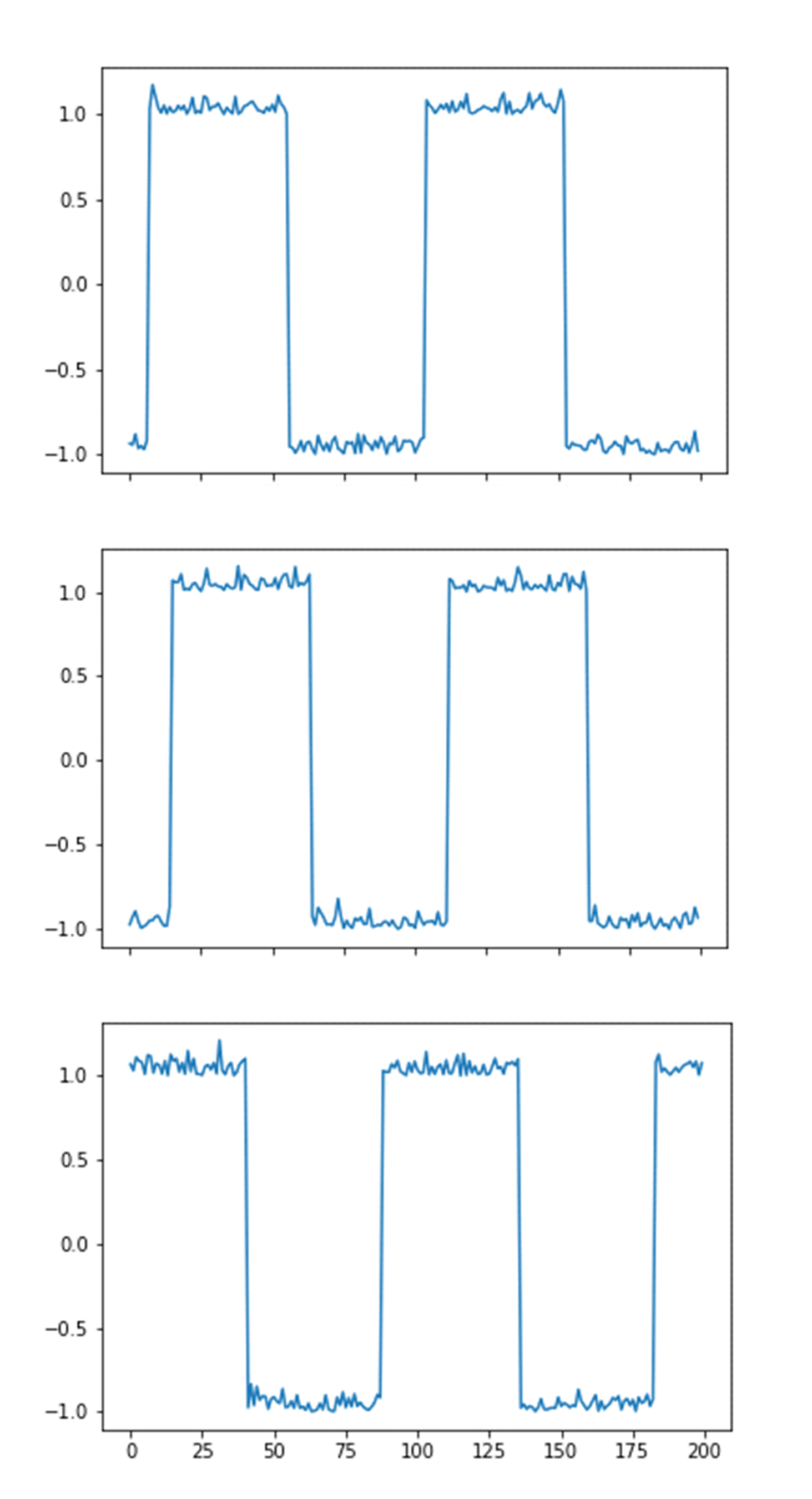}
    \end{minipage}}
    \subfigure[Triangle waves]{
    \begin{minipage}{0.31\columnwidth}
        \includegraphics[width=\columnwidth]{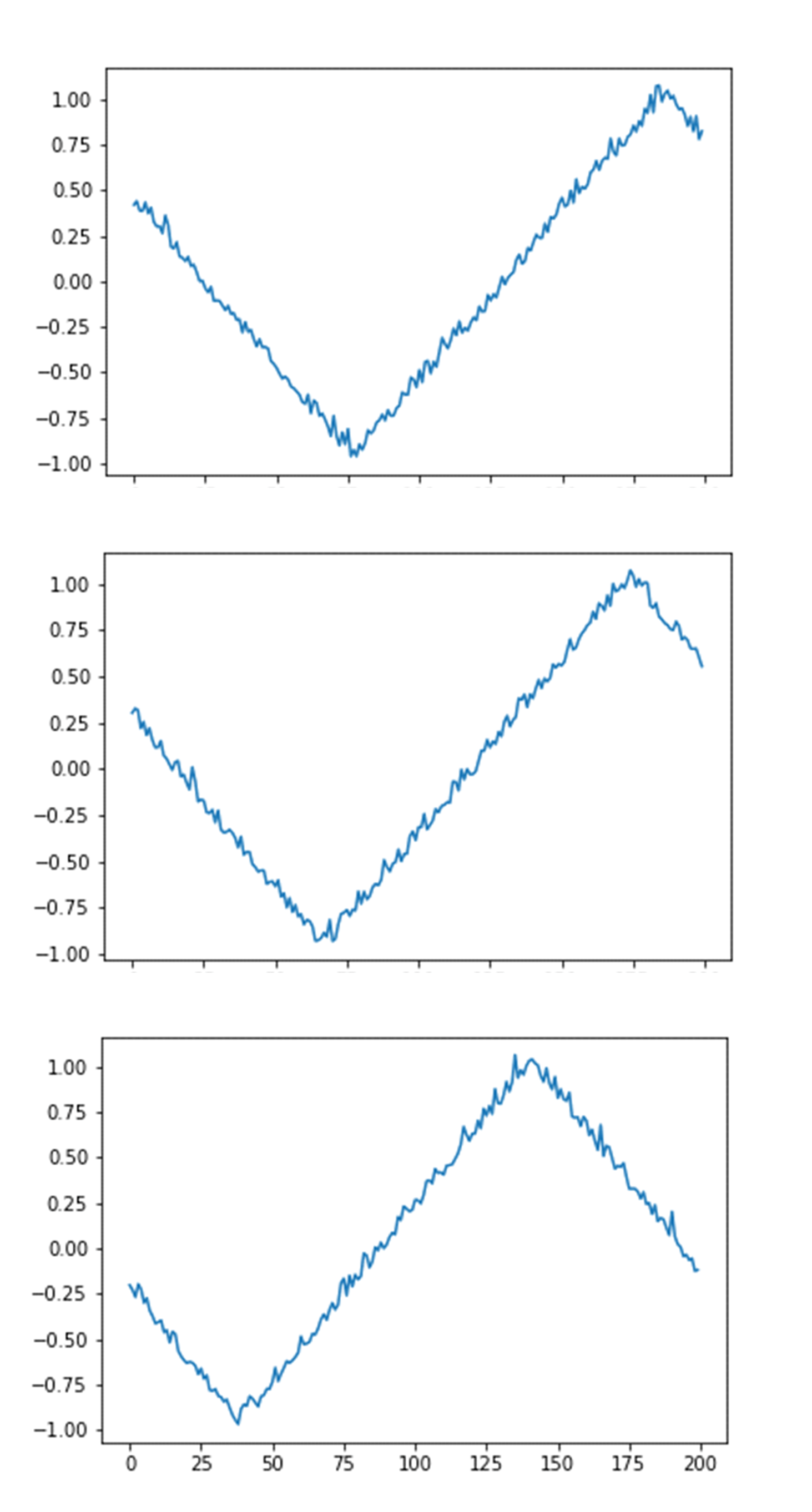}
    \end{minipage}}
    \caption{Exemplary samples from the synthetic time series clustering dataset}
\label{fig:exp_clustering}
\end{figure}

\begin{figure}[htbp]
\centering
    \subfigure[2D visualization result]{
    \begin{minipage}[!t]{0.48\columnwidth}
        \includegraphics[width=1.7in]{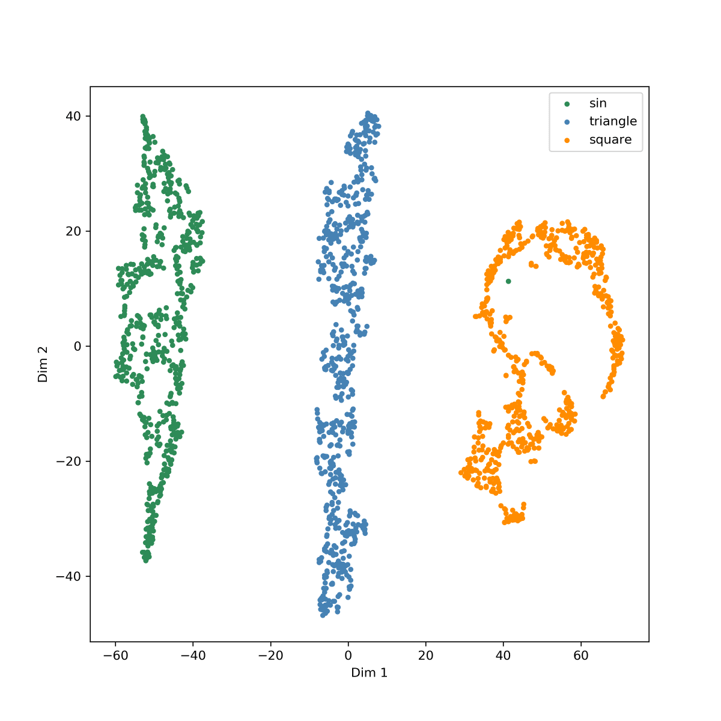}
    \end{minipage}}
    \subfigure[3D visualization result]{
    \begin{minipage}[!t]{0.48\columnwidth}
        \includegraphics[width=1.7in]{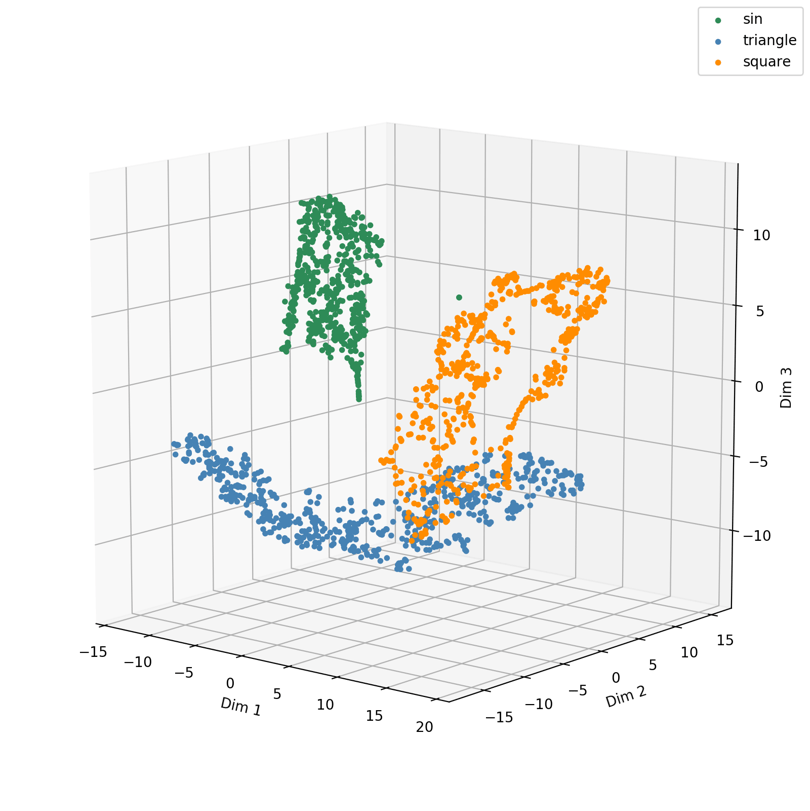}
    \end{minipage}}
    \caption{Clustering visualization results.
    Green, blue and orange symbols represent the vector embeddings of sin,
    triangle and square waves, respectively.}
\label{fig:result_clustering}
\end{figure}

This dataset consists of three types of time series, i.e., sine waves, square waves and triangle waves \textcolor{black}{(as shown in different columns of Figure \ref{fig:exp_clustering})}, we also pass these time series through an Additive white Gaussian noise (AWGN) channel \textcolor{black}{and introduce phase difference artificially}\textcolor{black}{(as shown in different rows of Figure \ref{fig:exp_clustering})} to make them more realistic. The vector embeddings in the latent space can be obtained for the time series through the ET-Net, as visualized in Figure \ref{fig:result_clustering} using the t-SNE algorithm. It is evident that we can easily cluster these time series in the latent space.

\begin{figure}[htbp]
\centering
    \subfigure[Type-1 noise]{
        \begin{minipage}{0.48\columnwidth}
        \includegraphics[width=\columnwidth]{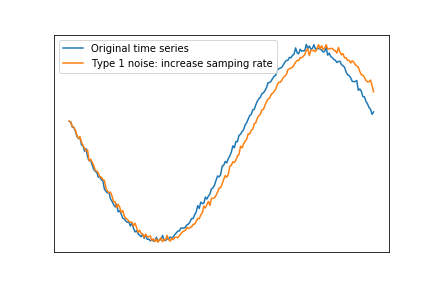}
        \end{minipage}}
    \subfigure[Type-2 noise]{
        \begin{minipage}{0.48\columnwidth}
        \includegraphics[width=\columnwidth]{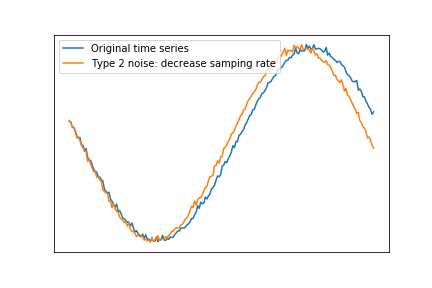}
        \end{minipage}}

    \subfigure[Type-3 noise]{
        \begin{minipage}{0.48\columnwidth}
        \includegraphics[width=\columnwidth]{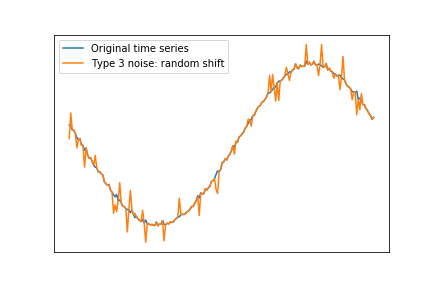}
        \end{minipage}}
    \subfigure[Type-4 noise]{
        \begin{minipage}{0.48\columnwidth}
        \includegraphics[width=\columnwidth]{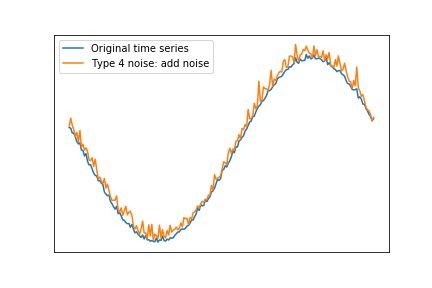}
        \end{minipage}}
\caption{Exemplary samples of different types of noise}
\label{fig:exp_noise}
\end{figure}

\textbf{Robustness against different types of noise:}
Four types of noise described in \cite{wang2013effectiveness} are considered in this experiment.
Type-1 and type-2 noise stand for increasing (Figure \ref{fig:exp_noise}(a)) and decreasing sampling rate (Figure \ref{fig:exp_noise}(b)), respectively.
Type-3 noise is the shift noise (Figure \ref{fig:exp_noise}(c)).
Type-4 noise refers to adding Gaussian noise to the entire time series (Figure \ref{fig:exp_noise}(d)).

\begin{figure}[htbp]
\centering
    \includegraphics[width=\columnwidth]{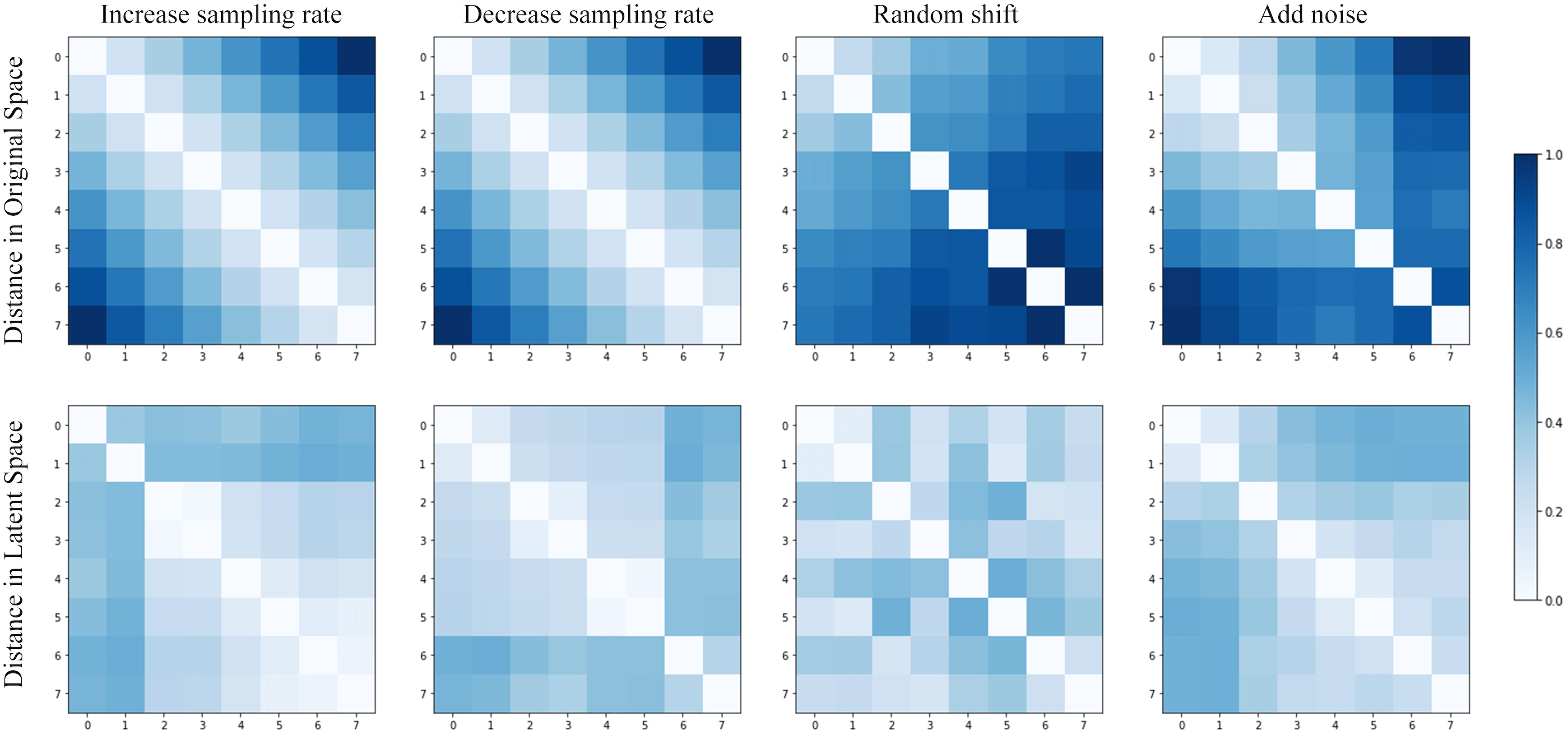}
    \caption{Normalized Euclidean distance matrix of a collection of time series in the
    original space (top row) and latent space (bottom row).
    The four columns are distance matrices when the time series contain
    different types of noise. }
\label{fig:result_noise}
\end{figure}

We then apply the four types of noise to a sine time series and compute the Euclidean distance between the original time series and the time series with noise in both original and latent spaces.
As shown in Figure \ref{fig:result_noise}, for all four types of noise, the Euclidean distance will increase quickly with the level of noise.
As a contrast, the proposed framework remains effective in the presence of all four types of noise and can mine the similarity between the original time series and the ones with noise.
As a matter of fact, as shown in Table \ref{table:noise}, it is the only method that remain robust against all types of noise.

\begin{table}[htbp]
\centering
\caption{Comparison results of time series similarity measures \cite{wang2013effectiveness}}
\label{table:noise}
\scalebox{1}{
\begin{tabular}{@{}ccccc@{}}
    \toprule
                           & ED        & DTW       & EDR       & ET-Net \\ \midrule
    \makecell[c]{Increase\\sampling\\rate} & Sensitive & Fair      & Sensitive & Robust      \\
    \makecell[c]{Decrease\\sampling\\rate} & Sensitive & Sensitive & Fair      & Robust      \\
    \makecell[c]{Random\\shift}           & Robust    & Robust    & Robust    & Robust      \\
    \makecell[c]{Add\\noise}              & Sensitive & Sensitive & Sensitive & Robust      \\ \bottomrule
\end{tabular}}
\end{table}

\subsection{Results on Real-world Datasets}
\subsubsection{Anomaly Detection}
\label{res_anomaly_dectection_real_world}
\textbf{Implementation Details:}
We conduct experiments on a work station with 48-core Xeon E5 CPUs, 64GB
RAM and 2 NVIDIA Titan V GPUs. TensorFlow 1.10 is employed to implement the
proposed ET-Net framework.

ET-Net contains four types of hyperparameters,
1) the number of branches in the W compression network;
2) the number of layers in the D compression network;
3) the number of neurons in each layer of RNN;
4) the number of mixture components in GMM.
LSTM cell and GRU cell are adopted in the W and D compression network respectively.
We use the Adam optimization algorithm \cite{kingma2014adam} to train
the proposed model, and the initial learning rate is set to $10^{-3}$.
The hyperparameter settings are listed in Table \ref{table:hyper_params}.
For other algorithms, we follow the default or recommended settings.

\begin{table}[htbp]
\centering
\caption{Hyperparameter settings}
\label{table:hyper_params}
\scalebox{1}{
    \begin{tabular}{@{}lcccc@{}}
    \toprule
    Dataset      & \makecell[c]{\# of\\branches} & \makecell[c]{\# of\\layers} & \makecell[c]{\# of\\neurons} & \makecell[c]{\# of GMM\\components} \\ \midrule
    \multicolumn{5}{c}{Anomaly Detection} \\ \midrule
    UNSW-IoT       & 3              & 2            & 18            & 4                    \\
    IoT23          & 3              & 2            & 18            & 4                    \\
    Cell traffic   & 3              & 4            & 18            & 1                    \\
    MedicalImages  & 3              & 2            & 8             & 3                    \\
    MoteStrain     & 3              & 2            & 8             & 1                    \\
    PowerCons      & 3              & 2            & 12            & 1                    \\
    SmoothSubspace & 3              & 2            & 8             & 1                    \\ \midrule
    \multicolumn{5}{c}{Clustering} \\ \midrule
    UNSW-IoT     & 3              & 2            & 12            & 3                    \\
    IoT23        & 3              & 4            & 18            & 8                    \\
    Cell traffic & 3              & 2            & 12            & 4                    \\ \bottomrule
    \end{tabular}}
\end{table}

\textbf{Effectiveness:}
The performance of ET-Net and other state-of-the-art methods on a total of seven real-world datasets are listed in Table \ref{table:auc}.
ET-Net-W and ET-Net-D represent W compression network with GMM and D compression network with GMM, respectively.
It is evident that the proposed ET-Net outperforms other competing methods considerably. It ranks first in five out of seven datasets, and ranks second in the remaining two datasets. In particular, for the cell traffic datasets, ET-Net outperforms the second best method by around 14\%. \textcolor{black}{Furthermore, ET-Net outperforms ET-Net-W and ET-Net-D thanks to the joint output of multiple networks based on multi-task learning.}

\begin{table*}
\centering
\caption{AUC scores}
\label{table:auc}

\scalebox{1}{
    \begin{tabular}{@{}lcccccccc@{}}
    \toprule
    Method           & UNSW-IoT        & Cell traffic    & IoT23           & MedicalImages   & MoteStrain      & PowerCons       & SmoothSubspace  \\ \midrule
    OCSVM            & 0.7947          & 0.3241          & 0.5000          & 0.5366          & 0.8017          & \underline{0.9827}    & \underline{0.9920}  \\
    LoF              & 0.8705          & 0.5940          & 0.6402          & 0.6689          & 0.5339          & 0.7630          & 0.9200          \\
    IF & 0.8586          & 0.4758          & 0.6544          & 0.5012          & 0.8959          & 0.9679          & \underline{0.9920}    \\
    DTW              & 0.8413          & 0.4865          & 0.6480          & 0.4508          & 0.8550          & 0.9642          & \textbf{1.0000} \\ \midrule
    GRU-AE           & 0.9099          & 0.4259          & 0.7430          & 0.6358          & 0.9122          & 0.9691          & 0.9840          \\
    Shared-SRNN     & 0.8279          & \underline{0.6944}    & 0.7657          & 0.4680          & \textbf{0.9397} & 0.9716          & 0.9720          \\
    DAGMM            & 0.8314          & 0.5000          & 0.7964          & 0.6473          & 0.5000          & 0.5333          & 0.7800          \\
    BeatGAN          & 0.7395          & 0.5029          & 0.8027          & 0.6318          & 0.9087          & 0.7864          & 0.7000          \\ \midrule
    ET-Net-W         & \underline{0.9356}    & 0.3982          & \textbf{0.8504} & \underline{0.7584}    & 0.7597          & 0.9382          & 0.9840         \\
    ET-Net-D         & 0.8696          & \textbf{0.8333} & 0.6819          & 0.5724          & 0.9016          & 0.9506          & \textbf{1.0000} \\
    ET-Net           & \textbf{0.9503} & \textbf{0.8333} & \underline{0.8289}    & \textbf{0.7608} & \underline{0.9270}    & \textbf{0.9975} & \textbf{1.0000} \\ \bottomrule
    \end{tabular}}
\end{table*}

In the previous study, we assume all the training dataset constitutes
non-anomalous time series. However, such an assumption does not always hold in practice, since a small portion of the training data might be anomalies.
As a remedy, we artificially inject anomalies into the training dataset to check whether we can still obtain an effective anomaly detector.
Table \ref{table:robustness} demonstrates that the proposed ET-Net architecture remains effective even in the presence 10\% of anomalies in the training dataset.

\begin{table}[htbp]
\caption{AUC scores with injected anomalies in training set}
\label{table:robustness}
\centering
\scalebox{1}{
    \begin{tabular}{@{}cclcc@{}}
    \multicolumn{2}{c}{UNSW-IoT} & \multicolumn{1}{c}{} & \multicolumn{2}{c}{Cell traffic} \\ \cmidrule(r){1-2} \cmidrule(l){4-5}
    Proportions & AUC Score &  & Proportions & AUC Score \\ \cmidrule(r){1-2} \cmidrule(l){4-5}
    0\% & 0.9503 &  & 0\% & 0.8333 \\
    5\% & 0.9177 &  & 5\% & 0.8229 \\
    10\% & 0.9071 &  & 10\% & 0.7779 \\ \cmidrule(r){1-2} \cmidrule(l){4-5}
    \textbf{loss} & \textbf{4.32\%} &  & \textbf{loss} & \textbf{5.54\%} \\ \cmidrule(r){1-2} \cmidrule(l){4-5}
    \end{tabular}}
\end{table}

\textbf{Robustness:}
Event-triggered sensors may generate traffic time series with different granularity, which also makes the traffic patterns extremely complex. We are naturally led to the following question: can a ET-Net trained by time series with one sampling interval be applied universally to time series with other intervals? To answer this question, we train a ET-Net model based on a training dataset with sampling interval equals to 60 seconds and then apply it to detect anomalous behaviors in time series with sampling intervals varying from 60 seconds to 120 seconds. As shown in Table \ref{table:sample_rate}, the model remains effective even the data granularity, demonstrating its robustness.

\begin{table}[htbp]
\centering
\caption{AUC scores on different sampling intervals}
\label{table:sample_rate}
\scalebox{1}{
    \begin{tabular}{@{}lcccc@{}}
    \toprule
    Sampling Intervals & 60sec           & 90sec           & 120sec          \\ \midrule
    OCSVM              & 0.7947          & 0.5010          & 0.5000          \\
    LoF                & 0.8705          & 0.8780          & 0.8664          \\
    IF   & 0.8586          & 0.8211          & 0.7939          \\
    DTW                & 0.8413          & 0.8511          & 0.8350          \\ \midrule
    GRU-AE             & \underline{0.9099}    & \underline{0.8903}    & \underline{0.8792}    \\
    Shared-SRNN       & 0.8279          & 0.7541          & 0.7317          \\
    DAGMM              & 0.8314          & 0.7591          & 0.7494          \\
    BeatGAN            & 0.7395          & 0.5837          & 0.6255          \\ \midrule
    ET-Net             & \textbf{0.9503} & \textbf{0.8909} & \textbf{0.9078} \\ \bottomrule
    \end{tabular}}
\end{table}

\textbf{Visualization and interpretability:}
From a perspective of latent space visualization, GMM models the distribution of normal samples using a GMM model, and forms a normal cluster in the latent space in the anomaly detection task, thus, the vector embedding that deviates from this distribution is deemed as an anomaly. A typical example is shown in Figure \ref{fig:result_anomaly}.

\begin{figure}[htbp]
\centering
    \includegraphics[width=\columnwidth]{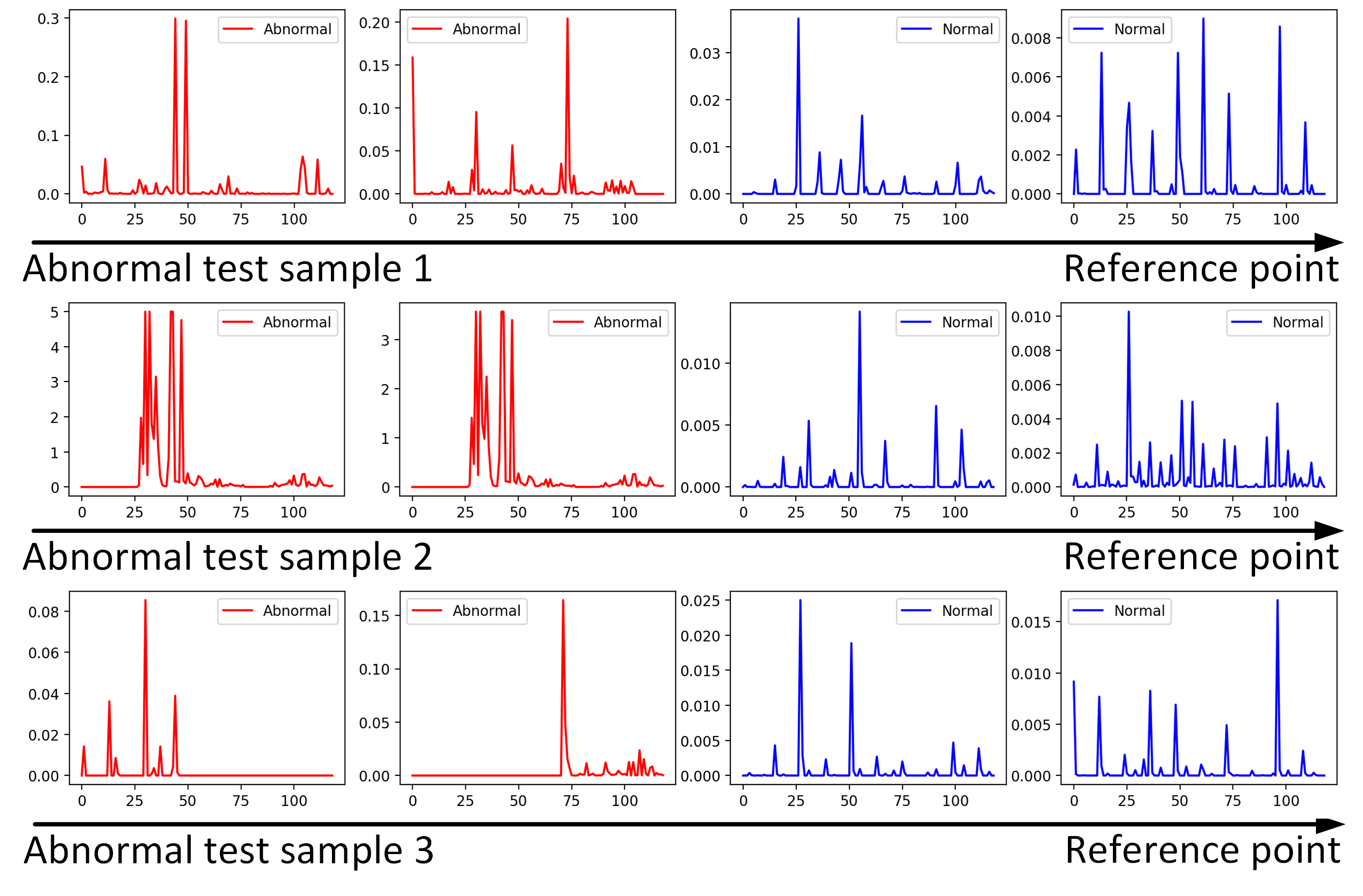}
    \caption{Exemplary test samples and corresponding reference samples from the UNSW-IoT dataset. Both anomalous and normal time series have been presented in these figures.}
\label{fig:visual_ts}
\end{figure}

\begin{figure*}[!t]
\centering
    \includegraphics[width=1.5\columnwidth]{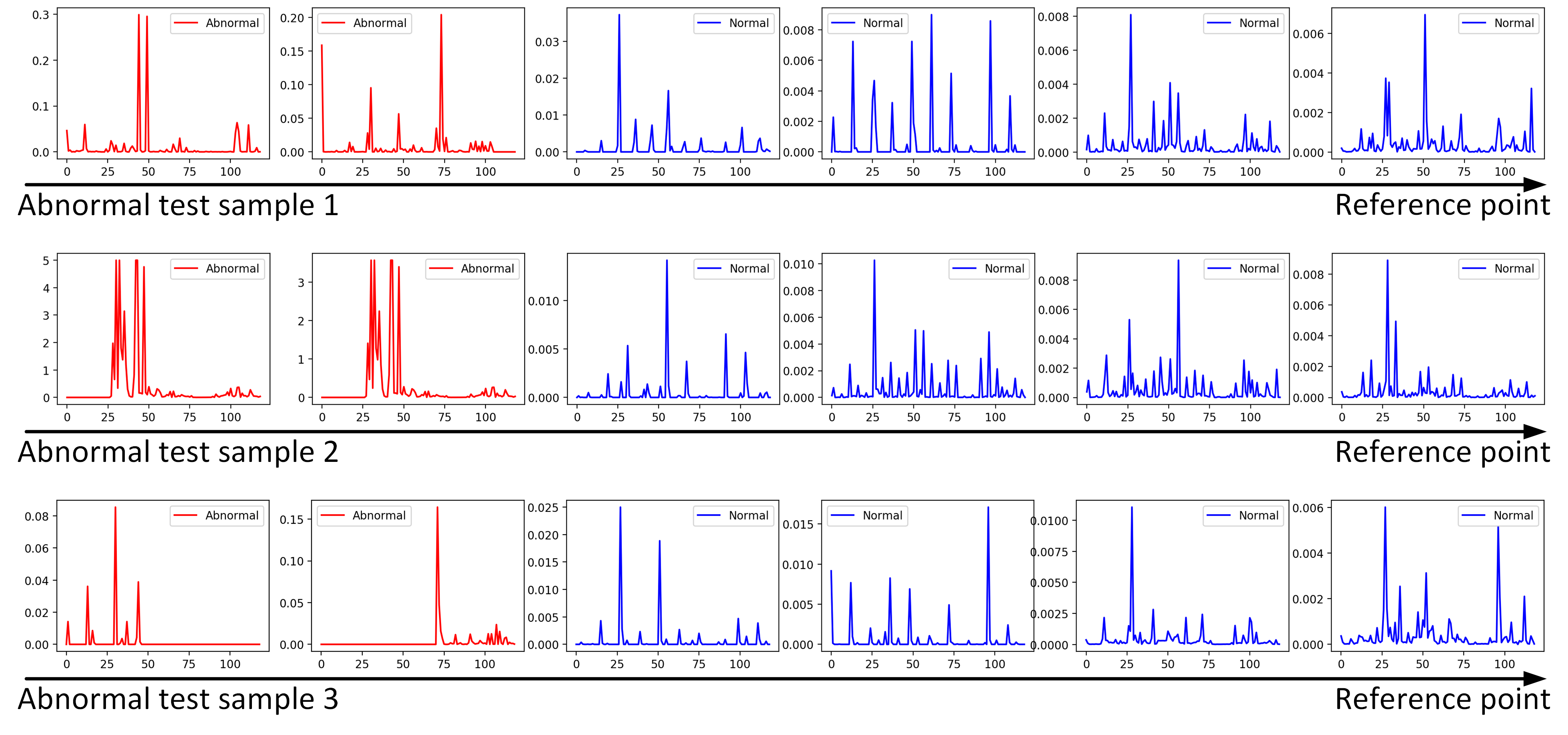}
    \caption{Exemplary test samples and corresponding reference samples from the UNSW-IoT dataset. The first column are the test samples, each subsequent column is closer to the center of the normal cluster.}
\label{fig:visual_ts_complete}
\end{figure*}

Based on the fact that normal samples are grouped into clusters in the latent space, we propose an example-based attribution method to explain the detected anomalies. Specifically, given a time series ${\underline{\bf{x}}}_a$ that is deemed as an anomaly, we draw a straight line in latent space from the low-dimensional representation of ${\underline{\bf{x}}}_a$, ${\underline{\bf{z}}}_a$ to the center of normal cluster ${\underline{\bf{z}}}_{cnt}$. We call this line in the latent space reference line hereafter. The visual comparisons among the anomaly time series and their corresponding reference time series help explain the difference between the anomalous times series and the normal ones.  
Figures in the first column in Figure \ref{fig:visual_ts} illustrate three representative abnormal time series, and the remaining time series are reference time series, where the second column of time series are the closest reference samples to the abnormal samples, and the third and fourth columns of samples are closer to ${\underline{\bf{z}}}_{cnt}$. See Figure \ref{fig:visual_ts_complete} for the complete figure. By observing these examples, we may extract semantic information that can help explain to users the difference between the anomalies and normal time series. 
\begin{itemize}
    \item Anomalous traffic time series may carry unusually high amount of traffic data compared with normal traffic time series, as given in the first two examples.
    \item Abnormal traffic time series may bear long and deep sleeping modes in which no traffic is transmitted.
\end{itemize}
Please notice that such semantic information extracted from these examples may be used to identify other anomalous time series as well.

\subsection{Time Series Clustering}
\textbf{Implementation Details:}
We conduct clustering experiments on three real datasets, including UNSW-IoT, IoT23 and cell traffic.
Hyperparameters used in the experiment are listed in Table \ref{table:hyper_params}.

\textbf{Effectiveness:}
Table \ref{table:result_cluster} lists the clustering performance of ET-Net and other state-of-the-art methods on three real-world datasets.
Two other methods have also been considered for comparison, including
1) AE+K-means in which the clustering is carried out over the latent space representations, which is obtained through the sequence autoencoder.
2) ET-Net+K-means in which the clustering is carried out over the vector embeddings obtained by the W and D compression network.
For both approaches, the same network hyperparameters as ET-Net
are adopted.

The results show that ET-Net outperforms all other state-of-the-art methods in two out of three datasets, and acquires the second-best in UNSW-IoT dataset. This substantiates the effectiveness of the ET-Net for unsupervised clustering.

\begin{table}[htbp]
\centering
\caption{NMI scores}
\label{table:result_cluster}
\scalebox{1}{
    \begin{tabular}{@{}lcccc@{}}
    \toprule
    Method         & UNSW-IoT        & IoT23           & Cell traffic    \\ \midrule
    K-means        & 0.0202          & 0.0399          & 0.0312          \\
    GMM            & 0.0000          & 0.0000          & 0.0107          \\
    K-means + DTW  & 0.5882          & 0.4860          & 0.0189          \\
    K-means + EDR  & 0.5046          & 0.3260          & 0.0318          \\ \midrule
    K-shape        & 0.6071          & 0.4558          & 0.0258          \\
    SPIRAL         & \textbf{0.9138} & 0.4390          & 0.0336          \\
    Autowarp       & 0.1002          & 0.1987          & 0.0281               \\
    DEC            & 0.0202          & 0.3091          & 0.0102          \\
    IDEC           & 0.0201          & 0.2569          & 0.0195          \\
    DTC            & 0.6117          & 0.1862          & 0.0000          \\ \midrule
    AE + K-means   & 0.7033          & 0.6268          & 0.0269          \\
    ET-Net + K-means & 0.6701          & 0.5659          & 0.0366          \\ \midrule
    ET-Net-W         & 0.4023          & \underline{0.6327}    & \underline{0.0542}    \\
    ET-Net-D         & 0.3734          & 0.2694          & 0.0193          \\
    ET-Net           & \underline{0.8304}    & \textbf{0.6753} & \textbf{0.0582} \\ \bottomrule
    \end{tabular}}
\end{table}

\section{Conclusion}
In this paper, we present ET-Net, a task-aware unsupervised deep learning approach to learn similarity metrics on event-triggered time series. Through extensive qualitative and quantitative studies, it is revealed that the proposed model can effectively capture the temporal dynamic of event-triggered time series. In addition, a single ET-Net model can be applied to time series with different time granularity with little performance degradation, which shows its robustness.



\appendices



\ifCLASSOPTIONcaptionsoff
  \newpage
\fi

\bibliographystyle{IEEEtran}
\bibliography{IEEEabrv,ijcai20}

%





\end{document}